\documentclass[11pt]{article}
\usepackage{coling2018}
\usepackage{times}
\usepackage{url}
\usepackage{latexsym}
\usepackage{booktabs}

\newcommand{\kawat}{\textbf{Ka}ta \textbf{W}ord \textbf{A}nalogy \textbf{T}ask}

\title{KaWAT: A Word Analogy Task Dataset for Indonesian}

\author{Kemal Kurniawan \\
  Kata Research Team \\
  Kata.ai \\
  Jakarta, Indonesia \\
  {\tt kemal@kata.ai} \\}

\date{}

\begin{document}
\maketitle
\begin{abstract}
  We introduced KaWAT (\kawat), a new word analogy task dataset for Indonesian.
  We evaluated on it several existing pretrained Indonesian word embeddings and
  embeddings trained on Indonesian online news corpus. We also tested them on
  two downstream tasks and found that pretrained word embeddings helped either
  by reducing the training epochs or yielding significant performance gains.
\end{abstract}

\section{Introduction}

Despite the existence of various Indonesian pretrained word embeddings, there
are no publicly available Indonesian analogy task datasets on which to evaluate
these embeddings. Consequently, it is unknown if Indonesian word embeddings
introduced in, e.g., \cite{al-rfou2013} and~\cite{grave2018}, capture
syntactic or semantic information as measured by analogy tasks. Also, such
embeddings are usually trained on Indonesian Wikipedia~\cite{al-rfou2013,bojanowski2017}
whose size is relatively small, approximately 60M tokens. Therefore, in this
work, we introduce KaWAT (\kawat), an Indonesian word analogy task dataset, and
new Indonesian word embeddings pretrained on 160M tokens of online news corpus.
We evaluated these embeddings on KaWAT, and also tested them on POS tagging and
text summarization as representatives of syntactic and semantic downstream task
respectively.

\section{Methodology}


We asked an Indonesian linguist to help build KaWAT based on English analogy
task datasets such as Google Word Analogy~\cite{mikolov2013} and
BATS~\cite{gladkova2016}. Following those works, we split the analogy tasks into
two categories, syntax and semantic. We included mostly morphological analogies
in the syntax category, leveraging the richness of Indonesian inflectional
morphology. For semantic, we included analogies such as antonyms, country
capitals and currencies, gender-specific words, measure words, and Indonesian
province capitals. In total, we have 15K syntactic and 19K semantic analogy
queries. KaWAT is available
online.\footnote{\url{https://github.com/kata-ai/kawat}}


One of the goals of this work is to evaluate and compare existing Indonesian
pretrained word embeddings. We used fastText pretrained embeddings introduced
in~\cite{bojanowski2017} and~\cite{grave2018}, which have been trained on
Indonesian Wikipedia and Indonesian Wikipedia plus Common Crawl data
respectively. We refer to them as Wiki/fastText and CC/fastText hereinafter. We
also used another two pretrained embeddings: polyglot embedding trained on
Indonesian Wikipedia~\cite{al-rfou2013} and NLPL embedding trained on the
Indonesian portion of CoNLL 2017 corpus~\cite{fares2017}.


For training our word embeddings, we used online news corpus obtained from
Tempo.\footnote{\url{https://www.tempo.co}} We used Tempo newspaper and magazine
articles up to year 2014. This corpus contains roughly 400K articles, 160M word
tokens, and 600K word types. To train the word embeddings, we experimented with
three algorithms: word2vec~\cite{mikolov2013a}, fastText~\cite{bojanowski2017},
and GloVe~\cite{pennington2014}. We refer to them henceforth as Tempo/word2vec,
Tempo/fastText, and Tempo/GloVe respectively. We used
\texttt{gensim}\footnote{\url{https://radimrehurek.com/gensim}} to run word2vec and
fastText and the original C implementation for
GloVe.\footnote{\url{https://github.com/stanfordnlp/GloVe}} For all three, we used
their default hyperparameters, i.e. no tuning was performed. Our three
embeddings are available
online.\footnote{\url{https://drive.google.com/open?id=1T9RmF0nHwN742aDkkQbjimpUCLgeVkhT}}


Evaluation on KaWAT was done using \texttt{gensim} with its
\texttt{KeyedVectors.most\_similar} method. Since the vocabularies of the word
embeddings are different, for a fair comparison, we first removed analogy
queries containing words that do not exist in any vocabulary. In other words, we
only kept queries whose words all exist in all vocabularies. After this process,
there were roughly 6K syntactic and 1.5K semantic queries. We performed
evaluation by computing 95\% confidence interval of the accuracy at rank 1 by
bootstrapping. Our implementation code is available
online.\footnote{\url{https://github.com/kata-ai/id-word2vec}}

\section{Results}

\subsection{Word analogy results}

We found that on syntactic analogies, Wiki/fastText achieved 2.7\% accuracy,
which significantly outperformed the others, even CC/fastText which has been
trained on a much larger corpus. Other embeddings performed poorly, mostly less
than 1\% of accuracy. The overall trend of low accuracy scores attests to the
difficulty of syntactic KaWAT analogies, making it suitable as benchmark for
future research.

On semantic analogies, Tempo/GloVe clearly outperformed the others with 20.42\%
accuracy, except Tempo/word2vec. Surprisingly, we found that
Tempo/fastText performed very poorly with less than 1\% accuracy, even worse than
Wiki/fastText which has been trained on a much smaller data. Overall, the
accuracies on semantic are also low, less than 25\%, which again attests to the
suitability of KaWAT as benchmark for future work.

\subsection{Downstream task results}

To check how useful these embeddings are for downstream tasks, we evaluated them
on POS tagging and text summarization task. For each task, we compared two
embeddings, which are the best off-the-shelf pretrained embedding and our
proposed embedding on the syntactic (for POS) and semantic (for summarization)
analogy task respectively.\footnote{We performed paired t-test and found
  Tempo/GloVe to be the best among our Tempo embeddings ($p < 0.05$).} We used
the same model and setting as~\cite{kurniawan2018} for POS tagging and
\cite{kurniawan2018b} for summarization. However, for computational reasons, we
tuned only the learning rate using grid search, and only used the first fold of
the summarization dataset. Our key finding from the POS tagging experiment is
that using the two embeddings did not yield significant gain on test $F_1$ score
compared with not using any pretrained embedding (around $97.23$). However, on
average, using Wiki/fastText resulted in 20\% fewer training epochs, compared
with only 4\% when using Tempo/GloVe. For the summarization experiment,
Tempo/GloVe was significantly better\footnote{As evidenced by the 95\%
  confidence interval reported by the ROUGE script.} than CC/fastText in
ROUGE-1 and ROUGE-L scores (66.63 and 65.93 respectively). The scores of using
CC/fastText was on par to those of not using any pretrained word embedding,
and we did not observe fewer training epochs when using pretrained word
embedding.

\section{Conclusion}

We introduced KaWAT, a new dataset for Indonesian word analogy task, and
evaluated several Indonesian pretrained word embeddings on it. We found that (1)
in general, accuracies on the analogy tasks were low, suggesting that
improvements for Indonesian word embeddings are still possible and KaWAT is hard
enough to be the benchmark dataset for that purpose, (2) on syntactic analogies,
embedding by~\cite{bojanowski2017} performed best and yielded 20\% fewer
training epochs when employed for POS tagging, and (3) on semantic analogies,
GloVe embedding trained on Tempo corpus performed best and produced significant
gains on ROUGE-1 and ROUGE-L scores when used for text summarization.

\section{Acknowledgment}

We thank Tempo for their support and access to their news and magazine corpora.
We also thank Rezka Leonandya and Fariz Ikhwantri for reviewing the earlier
version of this manuscript.

\bibliography{idword2vec}

\begin{thebibliography}{}

\bibitem[\protect\citename{{Al-Rfou} \bgroup et al.\egroup }2013]{al-rfou2013}
Rami {Al-Rfou}, Bryan Perozzi, and Steven Skiena.
\newblock 2013.
\newblock Polyglot: {{Distributed Word Representations}} for {{Multilingual
  NLP}}.
\newblock In {\em Proceedings of the {{Seventeenth Conference}} on
  {{Computational Natural Language Learning}}}, pages 183--192, Sofia,
  Bulgaria. {Association for Computational Linguistics}.

\bibitem[\protect\citename{Bojanowski \bgroup et al.\egroup
  }2017]{bojanowski2017}
Piotr Bojanowski, Edouard Grave, Armand Joulin, and Tomas Mikolov.
\newblock 2017.
\newblock Enriching word vectors with subword information.
\newblock {\em Transactions of the Association for Computational Linguistics},
  5:135--146.

\bibitem[\protect\citename{Fares \bgroup et al.\egroup }2017]{fares2017}
Murhaf Fares, Andrey Kutuzov, Stephan Oepen, and Erik Velldal.
\newblock 2017.
\newblock Word vectors, reuse, and replicability: {{Towards}} a community
  repository of large-text resources.
\newblock In {\em Proceedings of the 21st {{Nordic Conference}} on
  {{Computational Linguistics}}}, pages 271--276, Gothenburg, Sweden, May.
  {Link\"oping University Electronic Press, Link\"opings universitet / Language
  Technology Group, Department of Informatics, University of Oslo, Norway}.

\bibitem[\protect\citename{Gladkova \bgroup et al.\egroup }2016]{gladkova2016}
Anna Gladkova, Aleksandr Drozd, and Satoshi Matsuoka.
\newblock 2016.
\newblock Analogy-based detection of morphological and semantic relations with
  word embeddings: What works and what doesn't.
\newblock In {\em Proceedings of the {{NAACL Student Research Workshop}}},
  pages 8--15, San Diego, California. {Association for Computational
  Linguistics}.

\bibitem[\protect\citename{Grave \bgroup et al.\egroup }2018]{grave2018}
Edouard Grave, Piotr Bojanowski, Prakhar Gupta, Armand Joulin, and Tomas
  Mikolov.
\newblock 2018.
\newblock Learning {{Word Vectors}} for 157 {{Languages}}.
\newblock In {\em Proceedings of the {{Eleventh International Conference}} on
  {{Language Resources}} and {{Evaluation}} ({{LREC}}-2018)}, Miyazaki, Japan.
  {European Language Resource Association}.

\bibitem[\protect\citename{Kurniawan and Aji}2018]{kurniawan2018}
Kemal Kurniawan and Alham~Fikri Aji.
\newblock 2018.
\newblock Toward a {{Standardized}} and {{More Accurate Indonesian
  Part}}-of-{{Speech Tagging}}.
\newblock In {\em 2018 {{International Conference}} on {{Asian Language
  Processing}} ({{IALP}})}, pages 303--307, Bandung, Indonesia, November.
  {IEEE}.

\bibitem[\protect\citename{Kurniawan and Louvan}2018]{kurniawan2018b}
Kemal Kurniawan and Samuel Louvan.
\newblock 2018.
\newblock {{IndoSum}}: {{A New Benchmark Dataset}} for {{Indonesian Text
  Summarization}}.
\newblock In {\em 2018 {{International Conference}} on {{Asian Language
  Processing}} ({{IALP}})}, pages 215--220, Bandung, Indonesia, November.
  {IEEE}.

\bibitem[\protect\citename{Mikolov \bgroup et al.\egroup }2013a]{mikolov2013}
Tomas Mikolov, Kai Chen, Greg Corrado, and Jeffrey Dean.
\newblock 2013a.
\newblock Efficient estimation of word representations in vector space.
\newblock {\em arXiv preprint arXiv:1301.3781}.

\bibitem[\protect\citename{Mikolov \bgroup et al.\egroup }2013b]{mikolov2013a}
Tomas Mikolov, Ilya Sutskever, Kai Chen, Greg~S. Corrado, and Jeff Dean.
\newblock 2013b.
\newblock Distributed representations of words and phrases and their
  compositionality.
\newblock In {\em Advances in Neural Information Processing Systems}, pages
  3111--3119.

\bibitem[\protect\citename{Pennington \bgroup et al.\egroup
  }2014]{pennington2014}
Jeffrey Pennington, Richard Socher, and Christopher Manning.
\newblock 2014.
\newblock Glove: {{Global Vectors}} for {{Word Representation}}.
\newblock In {\em Proceedings of the 2014 {{Conference}} on {{Empirical
  Methods}} in {{Natural Language Processing}} ({{EMNLP}})}, pages 1532--1543,
  Doha, Qatar. {Association for Computational Linguistics}.

\end{thebibliography}
\bibliographystyle{acl}

\end{document}